\pdfoutput=1
\documentclass[12pt]{article}

\usepackage{times}
\usepackage{fullpage}
\usepackage{parskip}
\usepackage{authblk}

\usepackage{hyperref}
\usepackage{url}
\usepackage{mathtools}
\usepackage[square,sort,comma,numbers]{natbib}
\usepackage{cite}
\usepackage{amssymb}
\usepackage{graphicx}
\usepackage{tikz}
\usepackage{subcaption}

\title{Initialization Strategies of Spatio-Temporal Convolutional Neural Networks}

\author{Elman Mansimov, Nitish Srivastava and Ruslan Salakhutdinov\thanks{emansim, nitish, rsalakhu@cs.toronto.edu} }
\affil{Department of Computer Science, University of Toronto}
\date{}

\begin{document}

\maketitle

\begin{abstract}
We propose a new way of incorporating temporal information present in videos into Spatial Convolutional Neural Networks (ConvNets) trained on images, that avoids training Spatio-Temporal ConvNets from scratch. We describe several initializations of weights in 3D Convolutional Layers of Spatio-Temporal ConvNet using 2D Convolutional Weights learned from ImageNet. We show that it is important to initialize 3D Convolutional Weights judiciously in order to learn temporal representations of videos. We evaluate our methods on the UCF-101 dataset and demonstrate improvement over Spatial ConvNets.

\end{abstract}

\section{Introduction}

Recognizing the action performed in videos is one of the most challenging problems in computer vision. Compared to images that only contain spatial information about objects present in one shot (frame), videos represent a continuous flow of information that describes the physics of the world that we live in. In these settings, learning the underlying temporal representations of information present in videos becomes an important challenge faced in order to recognize the action.\par

\noindent
Despite the fact that GPUs are getting faster and getting more memory every year, training Spatio-Temporal ConvNets on large scale video datasets such as Sports-1M \citep{KarpathyCVPR14} and Facebook-380K \citep{C3D} from scratch is a very time consuming task, which requires nearly a month to complete training. At the same time, training Spatio-Temporal ConvNets on smaller datasets such as UCF-101 \citep{UCF101} and HMDB \citep{HMDB} leads to severe overfitting. Taking a ConvNet trained on ImageNet \citep{ImageNet} and fine-tuning it on individual video frames solves the problem of overfitting, but gives unsatisfying solution because this model doesn't learn temporal representations present in multiple frames.\par

\noindent
To tackle the above challenge, we propose several ways of initializing 3D convolutional weights, which learn temporal representations of videos, using 2D convolutional weights learned from images. This dramatically speeds up training of Spatio-Temporal ConvNets and reduces overfitting on relatively small video datasets. We found that in order to learn temporal representations present in videos and get improvements in accuracy, it is important to initialize weights in 3D convolutional layers judiciously. Otherwise, the Spatio-Temporal ConvNet remains stuck at the plateau that only extracts spatial information from videos and will not get improvements in accuracy. By appropriately initializing weights and using Composite LSTM that learned representations of video sequences \citep{unsupLSTM} trained on labelled examples present in UCF-101 and unlabelled examples from Sports-1M, we managed to nearly match the current best classification accuracy \citep{pyramid1} on RGB data extracted from UCF-101 which uses many additional tricks to improve their performance.\par

\section{Related Work}
Research in action recognition was mainly driven by advancements in object recognition in images, where those approaches were extended or adapted to deal with videos. Traditional shallow approaches consisted of three main stages. Firstly, sparse spatio-temporal interest points were detected in videos and extracted using Histogram of Oriented Gradients (HOG) \citep{HOG} and Histogram of Optical Flow (HOF) \citep{HOF}. Recently, Wang {\em et al.} \citep{Wang2013} have proposed dense trajectories which became state-of-the art hand-crafted features for action recognition.  Next, those features got combined into a fixed-sized vector description, such as Bag of Words (BoW) or Fisher Vectors (FV). Lastly, the standard classifier such as SVM was trained on BoW or FV represenation to distinguish among the classes of interest. This approach still has state-of-the art performance on UCF-101 and HMDB datasets.\par

\noindent
With the larger availability of labeled image data and advancements in parallel computing, Deep ConvNets \citep{Krizhevsky} overshadowed traditional approaches in extracting features of images. Motivated by those results, researchers \citep{KarpathyCVPR14} and \citep{C3D} have created large scaled labelled video datasets and trained Deep Spatio-Temporal ConvNets on them, which were originally proposed by Ji {\em et al.} \citep{conv3d}. Even though their models had access to temporal information presented in videos, they didn't perform better than simple Spatial ConvNet fine-tuned on image frames extracted from UCF-101 and HMDB datasets \citep{KarpathyCVPR14}, \citep{C3D}. Recently, Simonyan {\em et al.} \citep{Simonyan14b} showed that Deep Early Fusion Spatio-Temporal ConvNet trained on dense optical flow extracted from videos significantly boosts the performance of fine-tuned spatial ConvNet trained on images. Their approach nearly matched state-of-the-art performance on UCF-101 dataset. Additionally, \citep{pyramid1}, \citep{pyramid2}, \citep{pyramid3} recently showed that spatio-temporal pooling of Deep ConvNet features gives additional improvement in performance.\par

\section{Transforming Features}
\label{sec:transforming}

In this section we describe several ways of transforming 2D Convolutional Weights into 3D ones, without losing the spatial information learned by training Spatial ConvNet on ImageNet. Suppose that we have a 2D Convolutional Weight Matrix $W^{(2D)}$ derived from training a Spatial ConvNet on ImageNet. Our goal is to create a 3D Convolutional Weight Matrix $W^{(3D)}$ with temporal dimension $T$ using those weights. Each sub-matrix $W^{(3D)}_{t}$, $\forall t\in \{1,...,T\}$ has the same dimension as the original 2D Convolutional Weight Matrix.\par

\noindent
In a trained Spatial ConvNet, each layer expects an input from the layer below it to be within some specific range. It is important to initialize $W^{(3D)}$ in a certain way, so that the output of the 3D Convolutional Layer remains to be approximately in the same range as the output of the originally learned 2D Convolutional Layer. In order to enforce this constraint, the sum of all sub-matrices of $W^{(3D)}$ has to be equal to $W^{(2D)}$ at the initialization time, i.e. $\sum_{t=1}^{T}W^{(3D)}_{t}=W^{(2D)}$.\par

\noindent
Below we desribe four different kinds of initializations of $W^{(3D)}$, which we have considered.\par

\subsection{Initialization by Averaging. (IA) } 

Since the consecutive image frames in videos share similarities in appearance, we would expect $W^{(2D)}$ to extract very similar spatial representations from each consecutive image frame. Therefore, we can initialize each sub-matrix $W^{(3D)}_{t}$ to extract the same spatial representations present in its respective input feature map. This could be accomplished by diving each element of the 2D Convolutional weight matrix by the temporal size of the convolutional layer and setting all convolutional weights across temporal dimension of $W^{(3D)}$ to the resulting matrix. More formally, it can be expressed as:\par
$W^{(3D)}_{t}=\frac{W^{(2D)}}{T}$, $\forall t\in \{1,...,T\}$\par

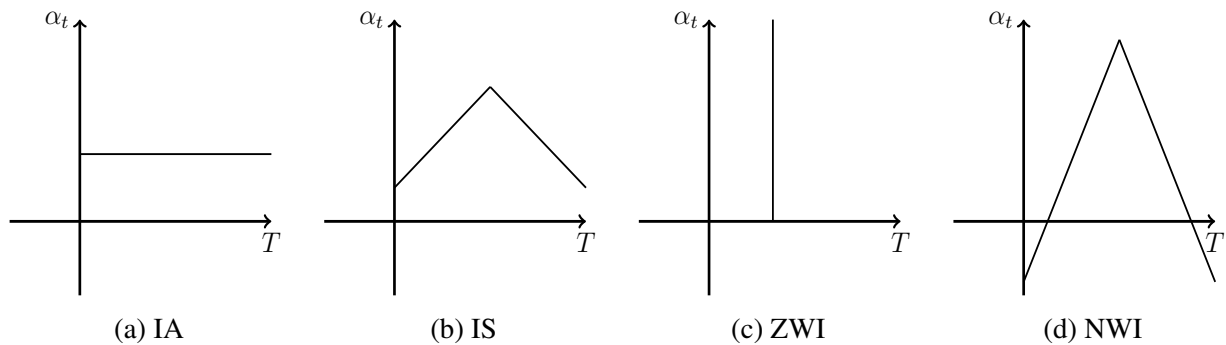
\begin{figure}
    \begin{subfigure}[t]{0.24\textwidth}
        \centering
        \resizebox{0.99\linewidth}{0.99\linewidth}{
			\begin{tikzpicture}
              \draw[->, very thick] (-1.1,0) -- (3,0) node[below] {$T$};
              \draw[->, very thick] (0,-1.1) -- (0,3) node[left]{$\alpha_{t}$};
              \draw[-, thick] (0,1)--(3,1);
			\end{tikzpicture}
        }
        \caption{IA}
        \label{fig:subfig8}
    \end{subfigure}
    \hfill
    \begin{subfigure}[t]{0.24\textwidth}
    \centering
        \resizebox{0.99\linewidth}{0.99\linewidth}{
			\begin{tikzpicture}
              \draw[->, very thick] (-1.1,0) -- (3,0) node[below] {$T$};
              \draw[->, very thick] (0,-1.1) -- (0,3) node[left]{$\alpha_{t}$};
			  \draw[-, thick] (0,0.5)--(1.5,2);
			  \draw[-, thick] (1.5,2)--(3,0.5);
			\end{tikzpicture}
        }
        \caption{IS}
        \label{fig:subfig9}
    \end{subfigure}
    \hfill
    \begin{subfigure}[t]{0.24\textwidth}
        \centering
        \resizebox{0.99\linewidth}{0.99\linewidth}{
			\begin{tikzpicture}
              \draw[->, very thick] (-1.1,0) -- (3,0) node[below] {$T$};
              \draw[->, very thick] (0,-1.1) -- (0,3) node[left]{$\alpha_{t}$};
			  \draw[-, thick] (1,0)--(1,3);
			\end{tikzpicture}
        }
        \caption{ZWI}
        \label{fig:subfig10}
    \end{subfigure}
    \hfill
	\begin{subfigure}[t]{0.24\textwidth}
        \centering
        \resizebox{0.99\linewidth}{0.99\linewidth}{
			\begin{tikzpicture}
              \draw[->, very thick] (-1.1,0) -- (3,0) node[below] {$T$};
              \draw[->, very thick] (0,-1.1) -- (0,3) node[left]{$\alpha_{t}$};
			  \draw[-, thick] (0,-0.9)--(1.5,2.7);
			  \draw[-, thick] (1.5,2.7)--(3,-0.9);
			\end{tikzpicture}
        }
        \caption{NWI}
        \label{fig:subfig11}
    \end{subfigure}

\caption{Example values of $\alpha_{t}$ for each initialization} 
\label{fig:subfig1.a.4}
\end{figure}

\subsection{Initialization by Scaling. (IS)}

This method can be viewed as a generalization to the method of initializing weights by averaging. Instead of dividing $W^{(2D)}$ by the same number $T$ and hence making each $W^{(3D)}_{t}$ equal to each other, we can induce some diversity by setting each $W^{(3D)}_{t}$ to $W^{(2D)}$ divided by some random constant. Any combination of the values of those constants could work, as long as the constraint described in section~\ref{sec:transforming} is satisfied. Therefore, $W^{(3D)}$ becomes:\par
\noindent

$W^{(3D)}_{t}=\alpha_{t} * W^{(2D)}$, where $\alpha_{t}>0$ and $\sum_{t=1}^{T}\alpha_{t}=1$\par
\noindent

\subsection {Zero Weight Initialization. (ZWI) }
It is natural to ask ourselves what would happen if we initialize one of $W^{(3D)}_{t}$ to $W^{(2D)}$ and initialize other sub-matrices to zero matrix. Would those sub-matrices initialized with zeros be able to learn to extract meaningful representations from the input even in spite of having limited training data ? Additionally, despite some differences in values, the distribution of weights is the same in each sub-matrix of $W^{(3D)}$ in the above initializations. Because of that the network might get stuck at the plateau which was reached by learning spatial representation on images and as a result it might not learn the temporal representation presented in multiple frames. Motivated by these points, we explored the following initialization technique, that can be expressed as:\par

\[
    W^{(3D)}= 
\begin{cases}
    W^{(2D)},& \text{if } t=1\\
    O, & \text{otherwise.}
\end{cases}
\]\par
\noindent

Note that, for any $1\leq t\leq T$, $W^{(3D)}_{t}$ could be initialized to $W^{(2D)}$, as long as all values of other sub-matrices are zero. However, in our experiments we found out that changing the order made no difference in resulting accuracy.

\subsection {Negative Weight Initialization. (NWI) }
We can extend the zero weight initialization by encouraging each sub-matrix $W^{(3D)}_{t}$, excluding the first one, to move even further from original values of $W^{(2D)}$. This could be accomplished by setting the values of sub-matrices $W^{(3D)}_{t}$, $2\leq t\leq T$ to negative signed values of $W^{(2D)}$ divided by some constant. This woud make the absolute values of $W^{(3D)}_{t}$, $t=1$ larger compared to other initializations. Again as in ZWI, changing the order of sub-matrices doesn't give any change in resulting accuracy. It can be expressed as following:\par 
\noindent

$W^{(3D)}_{t}=\alpha_{t} * W^{(2D)}$, where

\[
    \alpha_{t}= 
\begin{cases}
    \frac{2T-1}{T},& \text{if } t=1\\
    -\frac{1}{T}, & \text{otherwise.}
\end{cases}
\]\par

\subsection {Architecture Details.} 
The ConvNet with an architecture described here \url{https://github.com/TorontoDeepLearning/convnet/blob/master/examples/imagenet/CLS_net_20140801232522.pbtxt} was trained on ImageNet ILSVRC-2012. It yielded 13.5$\%$ top-5 error on ILSVRC-2012 validation set. In this study, we did not focus on using the best ImageNet model. Rather, we chose one that was convenient and easy to work with and focused on studying the relative impact of our proposed method. When fine-tuning, we removed the softmax layer and reduced the number of units in the last fully connected hidden layer from 4096 to 2048. Otherwise, the performance of the model dropped by 6-7 $\%$. We also used an aggressive dropout of 0.8 in both FC layers. The first two convolutional layers were transformed into 3D Convolutional layers with temporal size of 3, stride of 1 and temporal size of 2, stride of 1 respectively. The 3D pooling operation between those layers was perfromed on regions with temporal size of 2. 
The optimization in all convolutional layers started after 500 iterations in order to preserve good features learned by the model on ImageNet. Due to the limited size of the training set, other 3D Convolutional Networks with larger number of 3D Convolutional Layers performed worse than this model. \par

\section{Evaluation and Discussion}

\subsection {Dataset} The evaluation is performed on first split of UCF-101 dataset. UCF-101 contains 13.2K videos (25 frames/second) annotated into 101 classes, where each split contains 9.5K training videos.

\subsection {Effect of Different Initializations.}
The results of all transfer learning experiments are shown in Table 1. The results yielded many different conclusions.\par
\noindent
First, fine-tuning 3D ConvNet initialized by averaging or scaling yielded only small marginal improvement compared to fine-tuning 2D ConvNet. For the initialization by scaling, the best results were yielded by using $\alpha_{1}=\alpha_{3}=\frac{1}{4}$ and $\alpha_{2}=\frac{1}{2}$ for the first convolutional layer, and $\alpha_{1}=\alpha_{2}=\frac{1}{2}$ for the second convolutional layer. We think that since each sub-matrix across temporal dimension is initialized with similar weights which extract spatial features, the convolutional layer couldn't move beyond this plateau.\par 
\noindent
In contrast, despite the fact that all sub-matrices of $W^{(3D)}$, except for one, were initialized with zero matrix, the model did surprisingly better than models initialized using averaging or scaling. This could be explained by the fact that, at the initialization time the first sub-matrix which extracts spacial representations encouraged other sub-matrices to learn to extract temporal representations. Also, compared to other IA and IS, larger differences in values between first and other sub-matrices may have helped the model to learn to extract better temporal representations from the input. This might explain why the model initialized by Negative Weight Initialization yielded the best performance among them all.\par 
\noindent
Finally, by averaging our softmax class probabilities with softmax class probabilities of Composite LSTM, we managed to nearly match the current highest accuracy of deep learning based approaches on RGB data. We think that our performance can be further increased by using different tricks described in \citep{pyramid1} and using SVM fusion instead of averaging. The results are summarized in Table 2.\par
\noindent



\begin{table} [!ht]
\parbox{.45\linewidth}{
\centering
\begin{tabular}{|c c|}
 \hline
 \textbf{Model} & \textbf{Performance} \\ [0.5ex] 
 \hline
 Spatial ConvNet & 71.8 $\%$ \\ 
 \hline
 Init. by Averaging (IA) & 72.0 $\%$\\
 \hline
 Init. By Scaling (IS) & 72.4 $\%$\\
 \hline
 Zero Weight Init. (ZWI) & 73.3 $\%$\\
 \hline
 Negative Weight Init. (NWI) & \textbf{73.9} $\%$\\
 \hline
\end{tabular}
\caption{Comparisons of different initialization techniques in Spatio-Temporal ConvNet with simple Spatial ConvNet.}
}
\hfill
\parbox{.45\linewidth}{
\centering
\begin{tabular}{|c c|}
 \hline
 \textbf{Model} & \textbf{Performance} \\ [0.5ex] 
 \hline
 Spatial ConvNet \citep{Simonyan14b} & 72.8 $\%$ \\ 
 \hline
 C3D \citep{C3D} & 72.3 $\%$\\
 \hline
 C3D + fc6 \citep{C3D} & 76.4 $\%$\\
 \hline
 LRCN \citep{BerkeleyVideo} & 71.1 $\%$\\
 \hline
 Composite LSTM \citep{unsupLSTM} & 75.8 $\%$\\
 \hline
 NWI + Composite LSTM & \underline{78.3} $\%$\\
 \hline
 ConvNet Features \citep{pyramid1} & \textbf{79.0} $\%$\\
 \hline
\end{tabular}
\caption{Comparisons with other state-of-the-art neural networks based approaches on RGB data.}
}
\end{table}

\subsection{Two-Stream Spatio-Temporal ConvNets}

\begin{figure}
    \begin{subfigure}[t]{0.33\linewidth}
        \includegraphics[width=1\linewidth,natwidth=812,natheight=612]{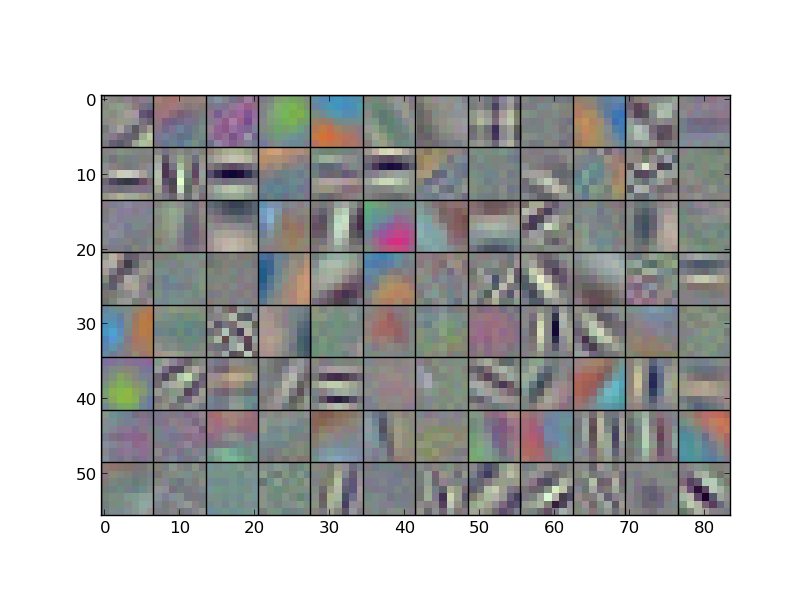}
    \end{subfigure}
    \begin{subfigure}[t]{0.33\linewidth}
        \includegraphics[width=1\linewidth,natwidth=812,natheight=612]{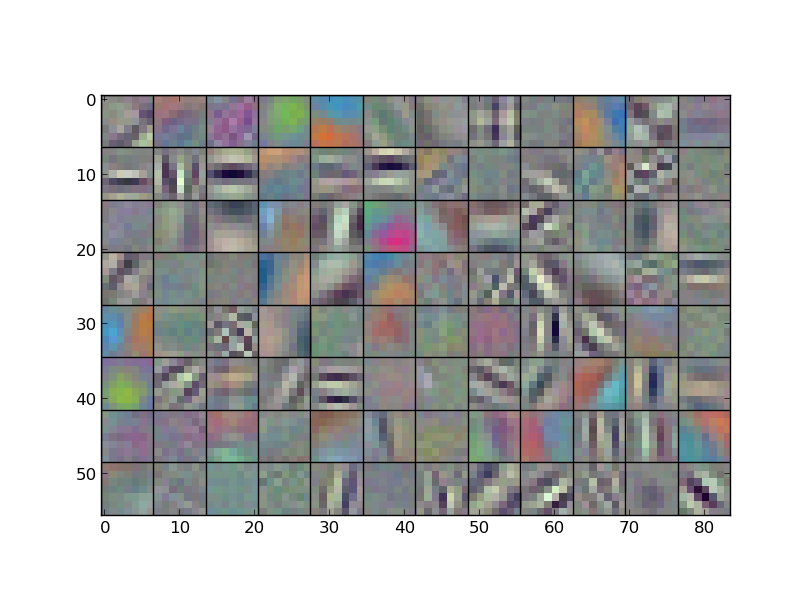}
        \caption{Initialization by Averaging}
    \end{subfigure}
    \begin{subfigure}[t]{0.33\linewidth}
        \includegraphics[width=1\linewidth,natwidth=812,natheight=612]{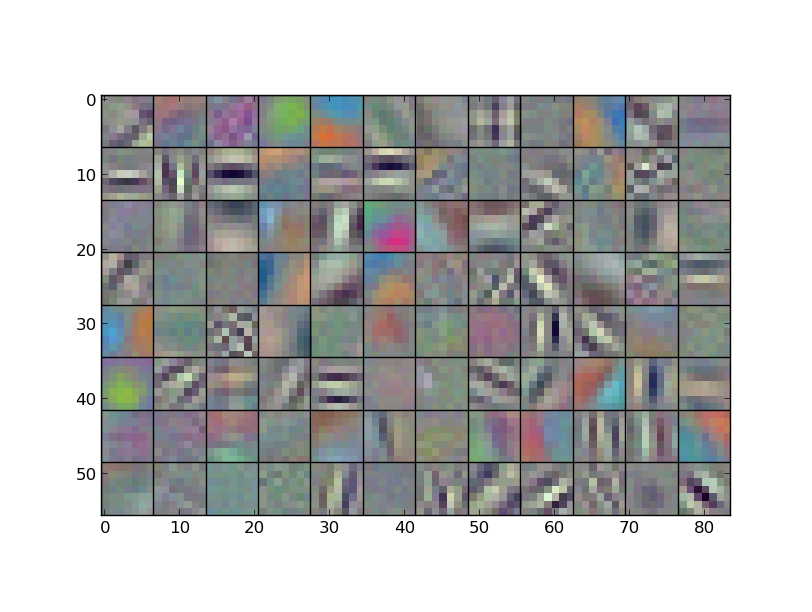}
    \end{subfigure}

    \begin{subfigure}[t]{0.33\linewidth}
        \includegraphics[width=1\linewidth,natwidth=812,natheight=612]{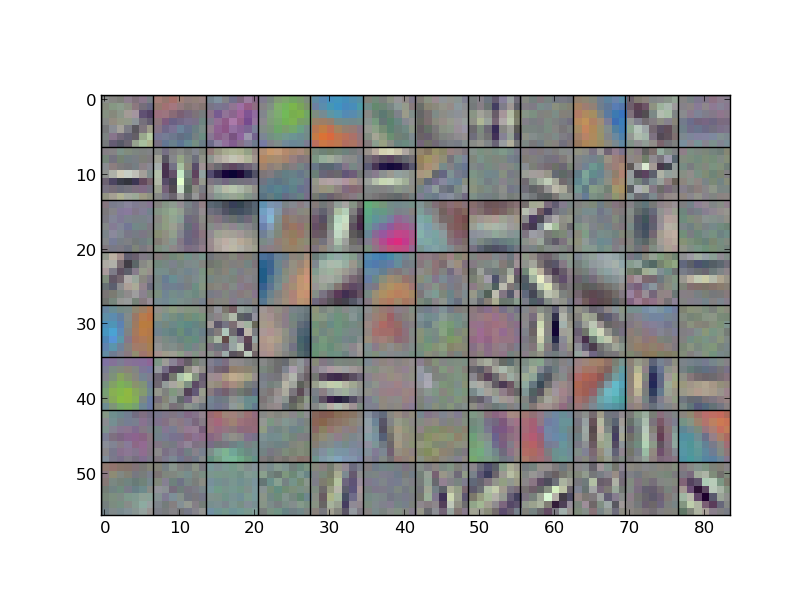}
    \end{subfigure}
    \begin{subfigure}[t]{0.33\linewidth}
        \includegraphics[width=1\linewidth,natwidth=812,natheight=612]{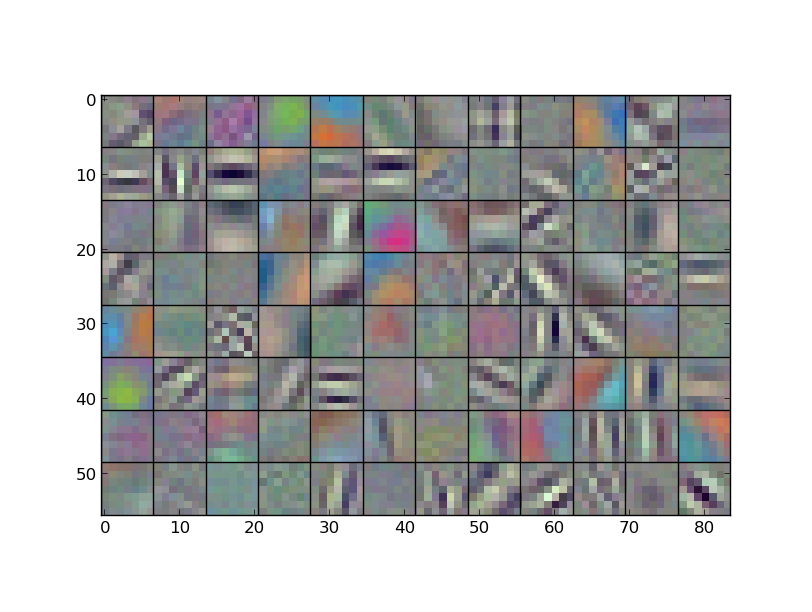}
        \caption{Initialization by Scaling}
    \end{subfigure}
    \begin{subfigure}[t]{0.33\linewidth}
        \includegraphics[width=1\linewidth,natwidth=812,natheight=612]{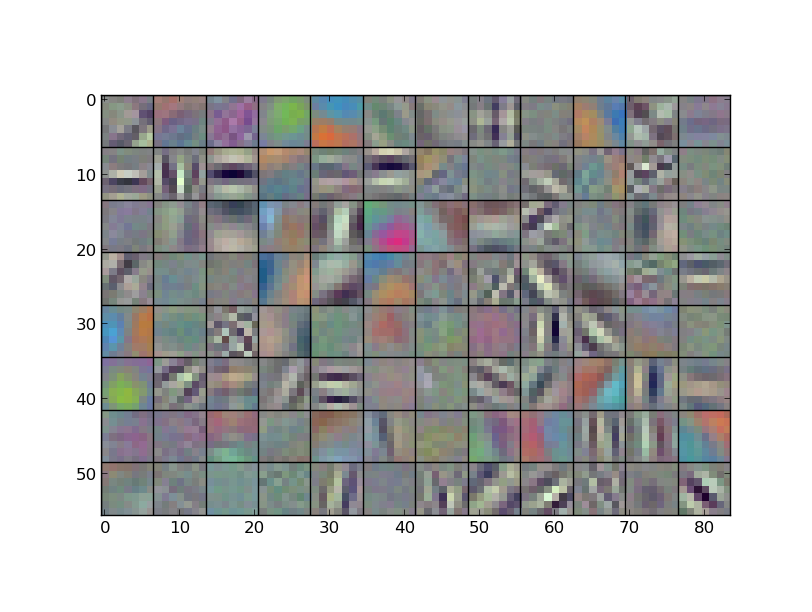}
    \end{subfigure}
    \begin{subfigure}[t]{0.33\linewidth}
        \includegraphics[width=1\linewidth,natwidth=812,natheight=612]{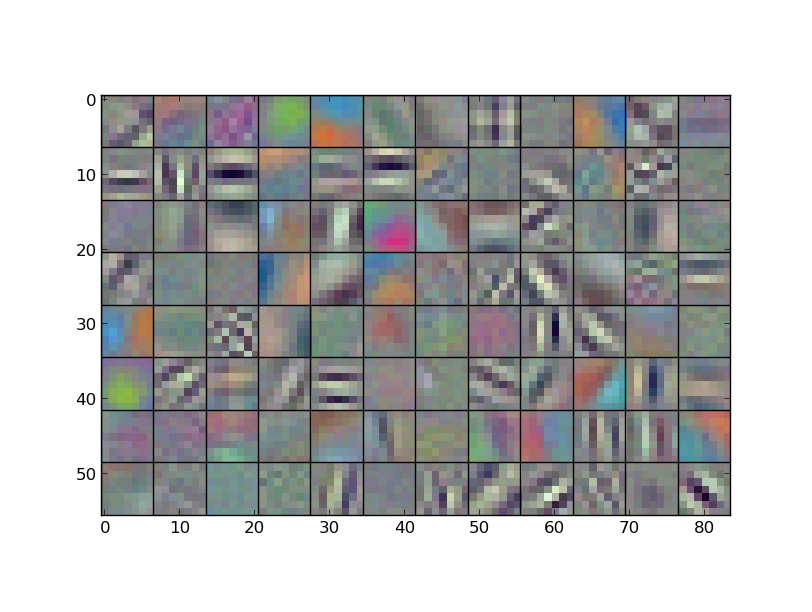}
    \end{subfigure}
    \begin{subfigure}[t]{0.33\linewidth}
        \includegraphics[width=1\linewidth,natwidth=812,natheight=612]{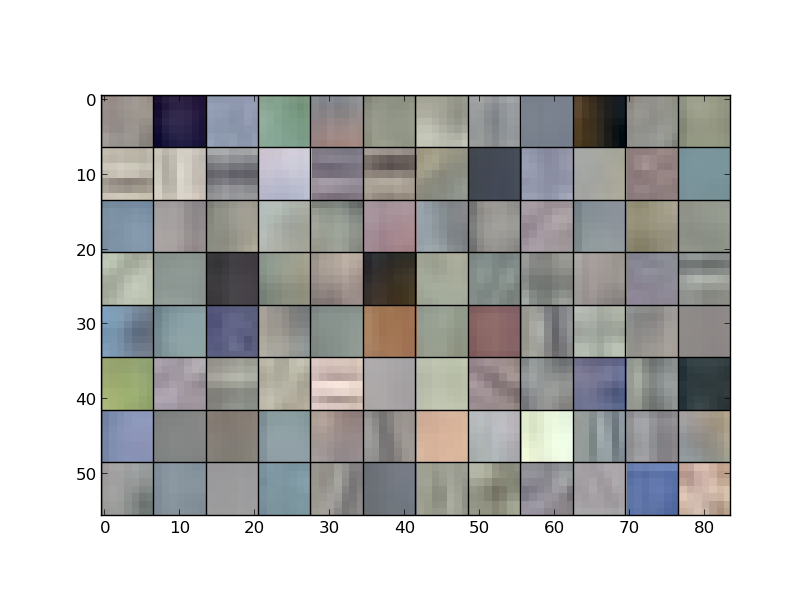}
        \caption{Zero Weight Initialization}
    \end{subfigure}
    \begin{subfigure}[t]{0.33\linewidth}
        \includegraphics[width=1\linewidth,natwidth=812,natheight=612]{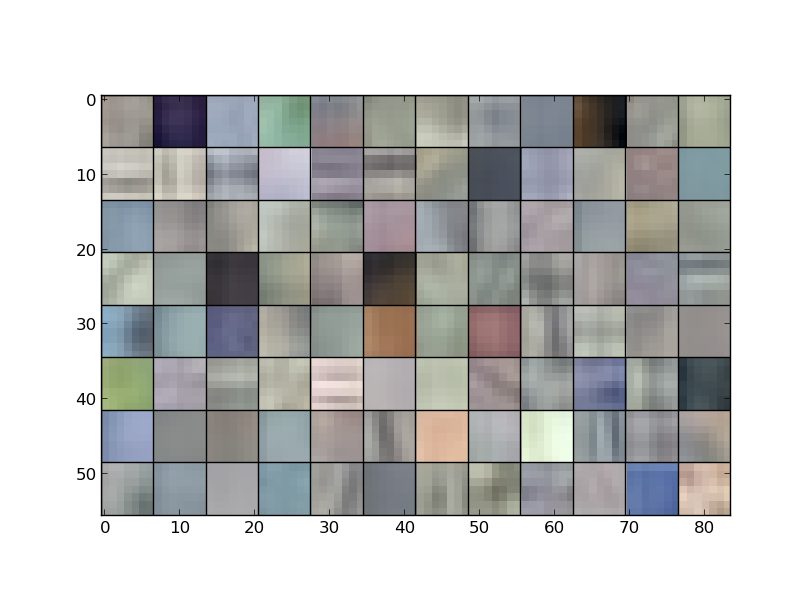}
    \end{subfigure}
    \begin{subfigure}[t]{0.33\linewidth}
        \includegraphics[width=1\linewidth,natwidth=812,natheight=612]{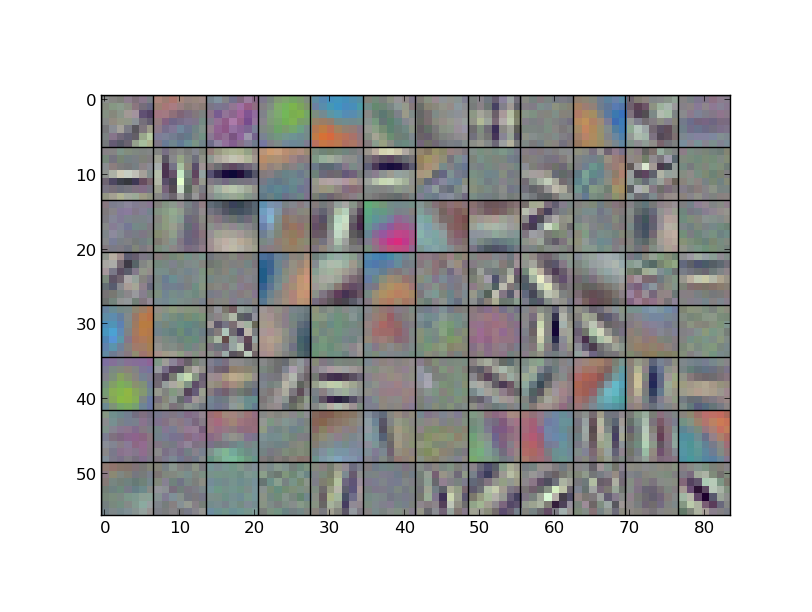}
    \end{subfigure}
    \begin{subfigure}[t]{0.33\linewidth}
        \includegraphics[width=1\linewidth,natwidth=812,natheight=612]{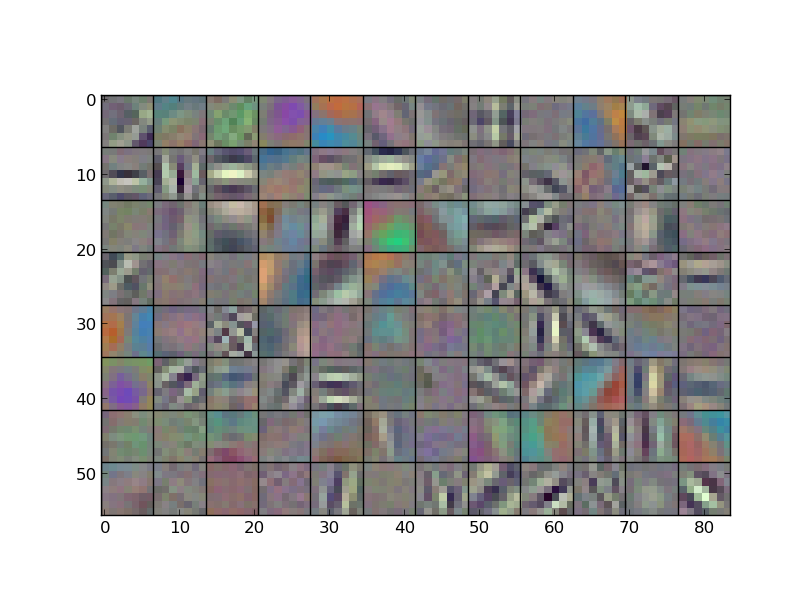}
        \caption{Negative Weight Initialization}
    \end{subfigure}
    \begin{subfigure}[t]{0.33\linewidth}
        \includegraphics[width=1\linewidth,natwidth=812,natheight=612]{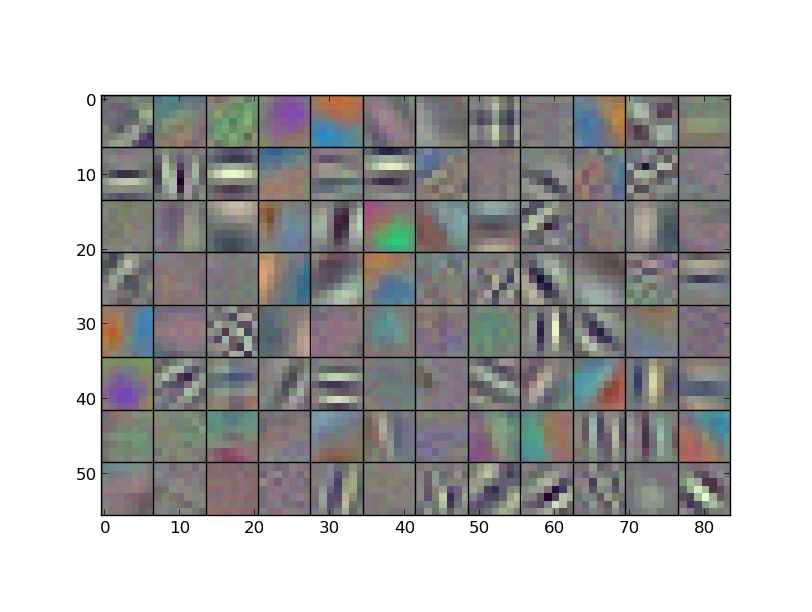}
    \end{subfigure}
\caption{3D Convolutional Kernels $W^{(3D)}=(W^{(3D)}_{1},W^{(3D)}_{2},W^{(3D)}_{3})$ learned by the first 3D Convolutional Layer initialized by IA, IS, ZWI and  NWI respectively. In Spatio-Temporal ConvNet initialized by averaging or scaling, $W^{(3D)}_{2}$ and $W^{(3D)}_{3}$ look very similar to $W^{(3D)}_{1}$. Whereas in ConvNet initialized by NWI, $W^{(3D)}_{2}$ and $W^{(3D)}_{3}$ look different from $W^{(3D)}_{1}$. Also, despite having limited labelled training video data, $W^{(3D)}_{2}$ and $W^{(3D)}_{3}$ look like edge and color blobs detectors in Spatio-Temporal ConvNet initialized by ZWI.} 
\end{figure}

We further ran experiments and evaluated complete two-stream model on first split, which combines RGB and Optical Flow based Spatio-Temporal ConvNets. Optical Flow based ConvNet, similar to the one proposed by \citep{Simonyan14b}, was trained on single frame optical flow and stacks of 10 optical flows. This gave an accuracy of 72.2$\%$ and 77.5$\%$ respectively. Additionally we trained a slow fusion version of this model, with an slow fusion setup same as in \citep{KarpathyCVPR14} and this model gave an accuracy of 79.3$\%$. We then combine the softmax scores of this optical flow based slow fusion model with the softmax scores of RGB based NWI + Composite LSTM model and obtain the accuracy of 85.3 $\%$ on UCF-101. Also, despite the smaller accuracy of early fusion model compared to its slow fusion version, the combination of optical flow based early fusion model with NWI + Composite LSTM gave the same accuracy of 85.3 $\%$. Esentially, despite better results of slow fusion model, it gives no additional performance improvement in final averaged scores. The comparison of our model with state-of-the-art action recognition models is summarized in Table 3.\par

\begin{table} [!ht]
\parbox{1.0\linewidth}{
\centering
\begin{tabular}{|c c|}
 \hline
 \textbf{Method} & \textbf{Performance} \\ [0.5ex] 
 \hline
 LRCN \citep{BerkeleyVideo} & 82.9 $\%$ \\ 
 \hline
 Two-Stream Convolutional Net \citep{Simonyan14b} (split 1) & 87.0 $\%$\\
 \hline
 C3D + fc6 + iDT \citep{C3D} & 86.7 $\%$\\
 \hline
 ConvNet Features + iDT \citep{pyramid1} & \textbf{89.7} $\%$\\
 \hline
 Multi-skip feature stacking \citep{MFS} & 89.1 $\%$\\
 \hline
 Composite LSTM Model \citep{unsupLSTM} (split 1) & 84.3 $\%$\\
 \hline
 Two-Stream Spatio-Temporal Convolutional Net (split 1) & \underline{85.3}  $\%$\\
 \hline
\end{tabular}
\caption{Comparisons of different action recognition models.}
}
\end{table}

\section {Conclusions}

We proposed a new way of creating Spatio-Temporal ConvNets, by incorporating temporal information into Spatial ConvNet trained on images. Despite having limited labelled video training data, our model outperformed Spatio-Temporal ConvNets trained on large scale labelled video datasets. It is interesting to see, whether initializing Spatio-Temporal ConvNet with one of the initialization techniques proposed here gives an improvement in performance on large scaled labelled video datasets. However, it is very challenging for us to process the datasets of such scale.

\small 
\bibliography{main}
\bibliographystyle{ieee}

\end{document}